\documentclass[fleqn,10pt]{wlscirep}
\usepackage[utf8]{inputenc}
\usepackage[T1]{fontenc}
\usepackage{amsmath}
\usepackage{amsfonts}
\usepackage{amssymb}
\usepackage{graphicx}
\usepackage{verbatim}
\usepackage[justification=justified]{caption}
\usepackage{url,hyperref}
\usepackage[capitalise]{cleveref}
\definecolor{tiffanyblue}{rgb}{0.04, 0.73, 0.71}
\definecolor{fuchsia}{rgb}{1.0, 0.0, 1.0}

\hypersetup{
    colorlinks=true,
    citecolor=tiffanyblue,
}

\newcommand{\figref}[1]{Fig.~\ref{#1}}

\newcommand{\ceqref}[1]{eq. \eqref{#1}}

\title{A framework for the emergence and analysis of language in social learning agents}

\author[1,2]{Tobias J. Wieczorek}
\author[1,3]{Tatjana Tchumatchenko{\href{https://orcid.org/0000-0001-9137-809X}{\includegraphics{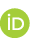}}}}
\author[3,4,\thanks{\,cwer1@uni-bonn.de},\#]{Carlos Wert Carvajal{\href{https://orcid.org/0000-0002-5340-8275}{\includegraphics{Figures/orcid_logo.eps}}}}
\author[1,\thanks{\,maximilian.eggl@uni-mainz.de},\#]{Maximilian F. Eggl{\href{https://orcid.org/0000-0001-5815-1045}{\includegraphics{Figures/orcid_logo.eps}}}}
\affil[1]{Institute of Physiological Chemistry, Johannes Gutenberg-University Mainz}
\affil[2]{Department of Physics, Technical University Darmstadt}
\affil[3]{Institute of Experimental Epileptology and Cognition Research, University of Bonn Medical Center}
\affil[4]{Max-Planck Institute for Brain Research, Frankfurt}
\affil[$\#$]{Equal contribution, corresponding authors}

\begin{abstract}
Artificial neural networks (ANNs) are increasingly used as research models, but questions remain about their generalizability and representational invariance. Biological neural networks under social constraints evolved to enable communicable representations, demonstrating generalization capabilities. This study proposes a communication protocol between cooperative agents to analyze the formation of individual and shared abstractions and their impact on task performance. This communication protocol aims to mimic language features by encoding high-dimensional information through low-dimensional representation. Using grid-world mazes and reinforcement learning, teacher ANNs pass a compressed message to a student ANN for better task completion. Through this, the student achieves a higher goal-finding rate and generalizes the goal location across task worlds. Further optimizing message content to maximize student reward improves information encoding, suggesting that an accurate representation in the space of messages requires bi-directional input. This highlights the role of language as a common representation between agents and its implications on generalization capabilities.

\textbf{Keywords:} Multi-agent learning, reinforcement learning, Q-learning, language emergence, dimensionality reduction, autoencoder, navigation, social learning

\end{abstract}

\begin{document}

\flushbottom
\maketitle

\thispagestyle{empty}

% \noindent Please note: Abbreviations should be introduced at the first mention in the main text – no abbreviations lists. Suggested structure of main text (not enforced) is provided below.

\section*{Introduction}

Studies considering the nature of task representations, whether linked to biological or artificial agents, have focused on those related to self-experience \cite{bengio2013representation,behrens2018cognitive}. However, abstractions are also essential for communication among individuals of the same species or group \cite{tomasello2009cultural}. Such social pressure implies that neural circuits may have evolved to produce internal representations that are not only useful for a given individual but may have co-evolved to maximize communication efficacy, which has been argued to be crucial in the development of cognition \cite{dunbar1998social,wilson2012social}. Previous studies on communication in reinforcement learning (RL) settings have focused mainly on performance consequences instead of the nature of the underlying neural representations \cite{giles2003learning,kasai2008learning,foerster2016learning}. Here, we focus on task-relevant communication or language in a broad sense \cite{hauser2002faculty}. Rather than modeling concrete symbols, grammar, or emulating human-like natural language, we assume it is a low-dimensional latent space within the messages that allow for shared representations of tasks or objects among individuals. Using this definition, we studied the co-evolution of task abstraction among different agents and addressed how sharing a common representation affects RL agent performance.

In this study, we posit that social aspects are crucial in providing task-efficient representations, particularly that there are fundamental characteristics of the task underlying the generalization of experiences among cooperative agents. The hypothesis that context and communication alter the task representation can be attributed to the introduction of language games \cite{wittgenstein1953philosophical}. Previous research in this direction focused on the conditions and constraints that would allow an artificial language to evolve and how similar this construction would be to human communication 
\cite{kirby1997learning,steels1997synthetic,cangelosi2002computer,wagner2003progress}. With the advent of deep learning and its application to RL, there has been a surge in papers that attempt to combine linguistic properties with deep neural network agents\cite{foerster2016learning,lazaridou2016multi,havrylov2017emergence,kottur2017natural,jaques2019social,rita2020lazimpa,kajic2020learning,ndousse2021emergent} (for a review see Oroojlooy and Hajinezhad (2022)\cite{oroojlooy2022review} or Lazaridou and Baroni (2020)\cite{lazaridou2020emergent}). This includes studies on multi-agent games where agents send and receive messages to perform tasks \cite{lazaridou2016multi}, tasks where the agents speak different languages and must learn to translate the other \cite{lee2017emergent} or multiple agents either compete or collaborate to develop representations of the tasks to form low-level policies \cite{sukhbaatar2018learning}.

In this work, we use RL to generate artificial agents who gather experiences while performing a navigational task \cite{sutton2018reinforcement}. This approach mimics the evolution of natural language, resulting from social and decision-making considerations \cite{seyfarth2014evolution}, in which individual abstractions emerge rather than being provided as predetermined labels, as done in a supervised learning fashion \cite{lee2017emergent,chaabouni2019anti,chaabouni2020compositionality}. Using these agents, we can then study the structure of the language embedding and how this affects the performance of agents.
The language embedding here consists of lower-dimensional representations of task information passed to students as labels. Notably, the second element of this architecture leads to insight into the most critical features that need to be represented for higher success at the task and generalization beyond those tasks. Finally, the architecture is flexible enough so that we can feed the representations of the student agent back through the language encoding and thus obtain a foundation for future study of languages that can naturally evolve and adapt to novel tasks beyond the scope of the original task set.

\section*{Results}

\section*{Model architecture}

To study the emergence of language between agents, we define two agents passing information to each other, a teacher and a student. Both of these agents are modeled as deep neural networks, whereby the teacher network is trained in an RL framework, and the student learns to interpret the instructions of the teacher \cite{sutton2018reinforcement,silver2018general,franccois2018introduction}. We used RL due to our interest in analyzing shared and generalizable abstractions arising from individual experiences and strategies instead of predetermined labels. Additionally, RL provides an intuitive and robust connection to neuroscience \cite{lee2012neural,sutton2018reinforcement}, which we aim to take advantage of to gain insight into the mechanisms and features of language emergence.

\begin{figure}
    \centering
    \includegraphics[width=.9\linewidth]{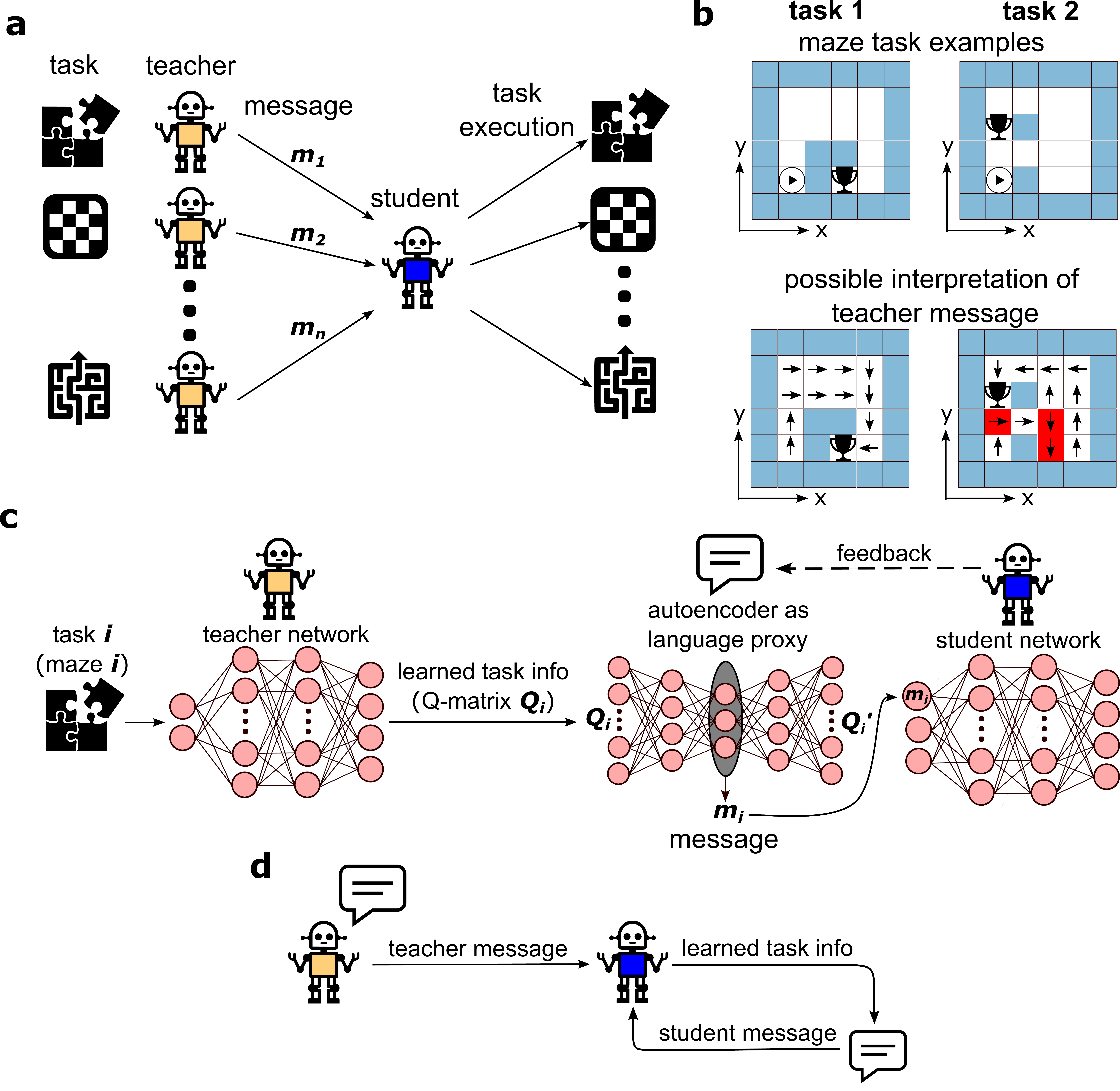}
    \caption{\textbf{Teacher-to-student communication model using a continuous compression of task solutions to low-dimensional message vectors.} \ {\bf a)} Model sketch which depicts a generalist student agent that is provided messages from teacher agents for various tasks. The student learns to decode these messages and then perform the relevant tasks.
      {\bf b)} \emph{Top)} Representative navigation tasks used to train and test agents to analyze the social learning framework. Beginning in the bottom left corner, the agents aim to reach the goal (trophy) in as few steps as possible while avoiding the walls (light blue squares). \emph{Bottom)} Overlayed example policies for those tasks learned by the teacher agents. The student needs to decode the encoded version of this information it receives. Messages may contain erroneous instructions or be misunderstood by the student (red squares). {\bf c)} Detailed communication architecture used in this study (the generalized framework allows for many tasks beyond the maze navigation setting). Task information (Q-matrices in our framework) is learned by teacher agents who then pass this information through a sparse autoencoder (language proxy), which generates the associated low-dimensional representations, $m_i$. The student ANN then receives the representation $m_{i}$ and learns to interpret it to solve task $i$. We also allow feedback from the student to propagate back to the language training for bi-directional communication (dashed line). {\bf d)} Schematic depicting the "closing-the-loop" architecture. Here the student is trained on a set of messages from an expert teacher. Once it is sufficiently competent, its task information is supplied to itself (after being passed through the language embedding), and the effect on performance is studied.}
    \label{fig:model_setup}
\end{figure}

In our setup, the teacher agent is presented with a task with complete access to its observations and rewards (\figref{fig:model_setup}). After a certain amount of training, the teacher will have obtained sufficient information to represent the task. The teacher network aims to produce a state-action value function or Q-matrix ($Q(s, a)$) of the task, which contains the expected return of state-action pairs, hence learning in a model-free and off-policy form. The student then aims to solve the same task but with additional information from the teacher, henceforth known as "message" (\figref{fig:model_setup}a). Thus, the student must learn and complete the task through their own observations and the message from the teacher. In our framework, we assume each teacher observes and learns from a single task and then passes a relevant task message – e.g., information derived from the Q-matrix – to the student. In that way, students can succeed on tasks they have yet to encounter by correctly interpreting the given information (\figref{fig:model_setup}b). 

The most relevant component of the architecture is the communication process. Natural language is a lower-dimensional representation of higher-dimensional concepts \cite{manning1999foundations}. When one individual speaks to another individual, high-dimensional descriptors – e.g., time, location, shape, context – of a concept in the brain of the sender are encoded into a low-dimensional vocabulary that is decoded back into a higher-dimensional and distributed representation in the brain of the receiver \cite{huth2016natural}. To mimic this interaction, we introduced a sparse autoencoder (SAE) \cite{ng2011sparse}, that takes the information from the teacher and produces a compressed message, $m$, that is passed to the student alongside the task. SAEs are also neural networks that consist of two parts, an encoder, and a decoder, and promote sparsity for the lower-dimensional representations. The encoder continuously projects the teacher network's output, $Q$, onto a message, $m$, which is a real-valued vector of length $K$. The decoder then uses this message to minimize the difference between its reconstruction $Q'$ and the true $Q$.

Furthermore, inspired by the sparse coding hypothesis \cite{huth2016natural}, we assume that the brain, and thus, by extension, language, is inherently sparsity-promoting \cite{manning1999foundations,olshausen2004sparse}. We implemented this by adding the norm of the message vector to the autoencoder loss, which follows the principle of least effort to guide our artificial communication closer to natural language (see \ceqref{eq:autoencoder_loss} in the \hyperref[sec:methods]{Methods}). The combination of one teacher, SAE, and student for an arbitrary task can be seen in \figref{fig:model_setup}c. Here, we utilized the $L^2$-norm of the message vector, which leads to a promotion of zeroes in the message and, therefore, less information that mimics sparsity.  

In this framework, we study a goal-directed navigational task in a grid-world maze (see \hyperref[sec:methods]{Methods}, \figref{fig:model_setup}b). We chose this relatively simple toy problem for the agents to learn due to its straightforward implementation – allowing us to focus on analyzing the message structure –, its usefulness in studying generalization and exploration strategies, and the possibility of extending it to more complex navigational settings \cite{sutton2018reinforcement}. We emphasize that the above architecture does not rely on a predetermined vocabulary for which the agents must assign meaning. Instead, the language evolves naturally from the task and the lower-dimensional encoding, mirroring natural language evolution.

The purpose of this study is two-fold; \emph{(i)} analyze the structure of the lower-dimensional representations generated by the trained language (which are lower-dimensional representations of our tasks), and \emph{(ii)} evaluate the performance of an agent who has learned to interpret a message coming from this embedding space.

\subsection*{The structure of the lower-dimensional message}
\begin{figure}
    \centering
    \includegraphics[width=.9\linewidth]{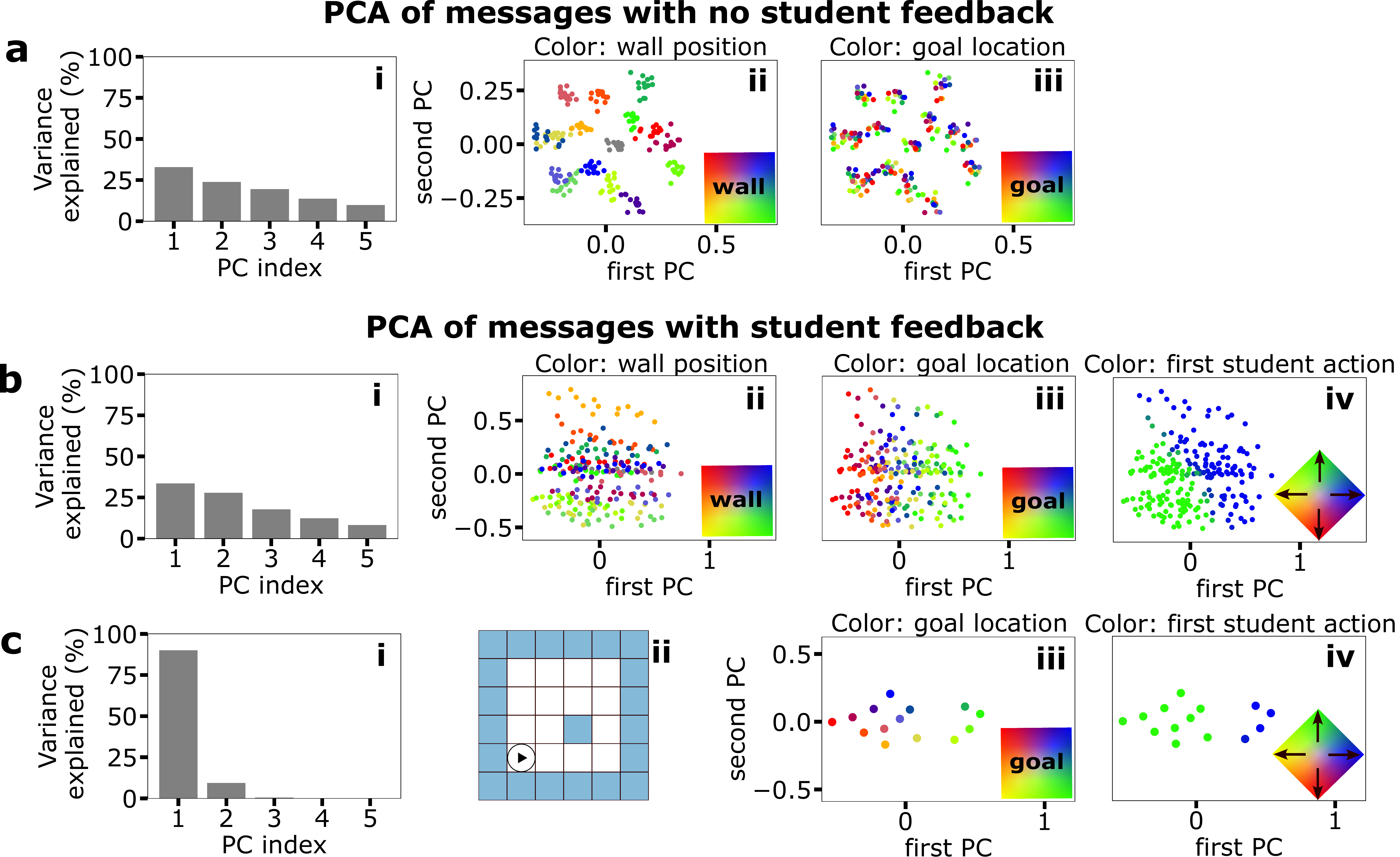}
    \caption{\textbf{Student feedback alters the language embedding according to a utility function.} \ {\bf a)} 
    Principal Component Analysis (PCA) of the lower-dimensional messages of size $K=5$ obtained from a language encoding without student feedback (\ceqref{eq:autoencoder_loss}) for all possible tasks in the $4 \times 4$ mazes with $\leq1$ walls (see \hyperref[sec:methods]{Methods} for a description of the tasks). {\it i)} Explained variance by principal component. {\it ii)-iii)} depicts the messages highlighted by the position of the single wall (gray refers to the maze with no walls) and by the position of the goal, respectively. {\bf b)} Result of the message encoding now including student feedback achieved by using \ceqref{eq:student_loss} for the loss function. {\it i)-iii)} depict the same concepts as in \emph{a)}. {\it iv)} shows messages highlighted by the preferred first student action (step up or right). {\bf c)} PCA of the messages with student feedback from an example grid-world (depicted in {\it ii)}).}
    \label{fig:nonlinear message space}
\end{figure}

We trained a set of teachers to solve each one maze task with a specific goal location and wall setting. As mentioned above, we use the trained language to embed the Q-matrices into a lower-dimensional space - firstly, considering a language created without feedback by the student. The resulting latent space shows wall positions as the most prominent dimension in the lower-dimensional representations (\figref{fig:nonlinear message space}a(ii)), with goal locations being a secondary feature of the variability (\figref{fig:nonlinear message space}a(iii)). We note that when we used linear activations or singular value decomposition for the language encoding, we did not reproduce this clear grouping (cf. \figref{fig:all linear message space}). Given that the language training without student feedback only relies on the reconstruction of the Q-matrix and regularization of the message space (\ceqref{eq:autoencoder_loss}), the most pertinent information is selected regardless of whether this information is useful for the student. 

While direct labeling of the tasks by such dimensions may help the student solve trained tasks, the average performance concerning trained tasks and generalization is significantly lower than when student feedback helps shape the language (see \figref{fig:student performance frozen1}). Furthermore, this interaction is purely one-directional and does not reflect the natural emergence of language, which is a back-and-forth between the receiver and sender. Therefore, we introduced student feedback into the message structure to encourage this natural evolution of language. Such feedback is implemented by including and maximizing the probability of the student finding the goal in the language training. This translates to a compound autoencoder loss function of the form
\begin{equation}\label{eq:student_loss}
    \mathcal{L}_{\textrm{SAE, feedback}}=\mathcal{L}_{\textrm{SAE}}+\zeta\mathcal{L}_{\textrm{goal finding}}=\mathcal{L}_{\textrm{reconstruction}}+\mathcal{L}_{\textrm{sparsity}}+\zeta\mathcal{L}_{\textrm{goal finding}},
\end{equation}
where the $\mathcal{L}_{\textrm{goal finding}}$ is defined by \ceqref{eq:goal_finding_loss} in the  \hyperref[sec:methods]{Methods} and $\zeta$ is a tunable hyperparameter. After each trial of the student, the language is updated to \emph{(i)} maintain the reconstruction of the information, \emph{(ii)} promote sparsity of the message, and \emph{(iii)} increase the success rate of the student given the message. 

Notably, the latent structure of the language space significantly changes through this reward-maximizing term (\figref{fig:nonlinear message space}b(ii)-(iv)). Even if the variance distribution remains similar (compare \figref{fig:nonlinear message space}a(i), \figref{fig:nonlinear message space}b(i)), task settings are no longer clustered in the latent space, but instead form a more continuous gradient when marked by wall position (\figref{fig:nonlinear message space}b(ii)) or goal location (\figref{fig:nonlinear message space}b(iii)). Therefore, the feedback changes the lower-dimensional task representations so that the student obtains more information on where to go, i.e., the policy, rather than the actual composition of the state space. We note some overlap in the middle of the cluster when marking the tasks by goal location; here, the policy differences are negligible as there might be two competing policies that are equally optimal. This focus on policy is additionally emphasized by the variability along the initial action of the student (\figref{fig:nonlinear message space}b(iv)), where a clear split between the two choices of going right or up can be observed. By providing this policy label, language moves away from providing maze labels and towards a framework that can generalize to tasks the student has not seen before. Table \ref{tab:variances_nonlinear} shows the changes in explained variability by wall position and goal location in both languages without and with student feedback. Notably, the message variability between groups of goal locations (see \hyperref[sec:methods]{Methods}) rises when the utility constraint is introduced, marking the increased importance of describing the goal location accurately in the language.

We can extend this analysis to understand the student feedback representation of the different goal locations for a single maze, where more than 80\% of the variance is explained by a single principal component (\figref{fig:nonlinear message space}c(i)). Geometric stucture (\figref{fig:nonlinear message space}c(iii)) and action selectivity (\figref{fig:nonlinear message space}c(iv)) are well represented in the embedding, the former indicating that language is performing a simple linear transformation of the geometric shape of the maze. We hypothesize that such information hierarchy benefits overall learning and generalization. This can also be seen by the performance exhibited by a novel student with a transferred language – i.e., frozen autoencoder – that was trained with the reward loss feedback (\figref{fig:student performance frozen2}). We note that these results hold independently of the activation function (\figref{fig:all linear message space}, \figref{fig:autoenc linear message space}).

Interestingly, the addition of student feedback reduced the overall reconstruction error of the message space (\figref{fig:student performance and loss}a-c, \ceqref{eq:autoencoder_loss}). This may suggest that the reconstruction of the teacher Q-matrix benefits from including features guided by utility and transmissibility criteria. Nevertheless, this comes at the cost of lower sparsity (\figref{fig:student performance and loss}c), mirroring the effect of natural language: communication aims to transmit the most sparse message, allowing for the best reconstruction of the underlying idea. Overall, we find these three items achieve a similar level of compound loss in both feedback and non-feedback (\figref{fig:student performance and loss}d).

\begin{table}[h!]
    \centering
    \begin{tabular}{l c c c c}
        \toprule
        \textbf{Message grouping} & \textbf{$\textrm{Var}_{\textrm{within}}(X)$} & \textbf {$\textrm{Var}_{\textrm{between}}(X)$} & $\beta$ & F-value \\
        \midrule
        By wall position (\figref{fig:nonlinear message space}a(ii)) & 2.88 & 20.18 & 0.875 & 97.54$^*$ \\
        By goal location (\figref{fig:nonlinear message space}a(iii)) & 19.99 & 3.07 & 0.133 & 2.14$^*$\\
        By wall position with student feedback (\figref{fig:nonlinear message space}b(ii)) & 20.06 & 38.07 & 0.655 & 26.44$^*$ \\
        By goal location with student feedback (\figref{fig:nonlinear message space}b(iii)) & 38.26 & 19.87 & 0.342 & 7.24$^*$ \\
        \bottomrule
    \end{tabular}
    \caption{Analysis of variance for world groups and goal groups in the message spaces from \figref{fig:nonlinear message space}. Statistical analysis is described in the \hyperref[sec:methods]{Methods}. \,$^*$ refers to a significant difference in group means with significance level set at $p=0.05$.}
    \label{tab:variances_nonlinear}
\end{table}

\subsection*{The effect of the message on student performance}

\begin{figure}
    \centering
    \includegraphics[width=.9\linewidth]{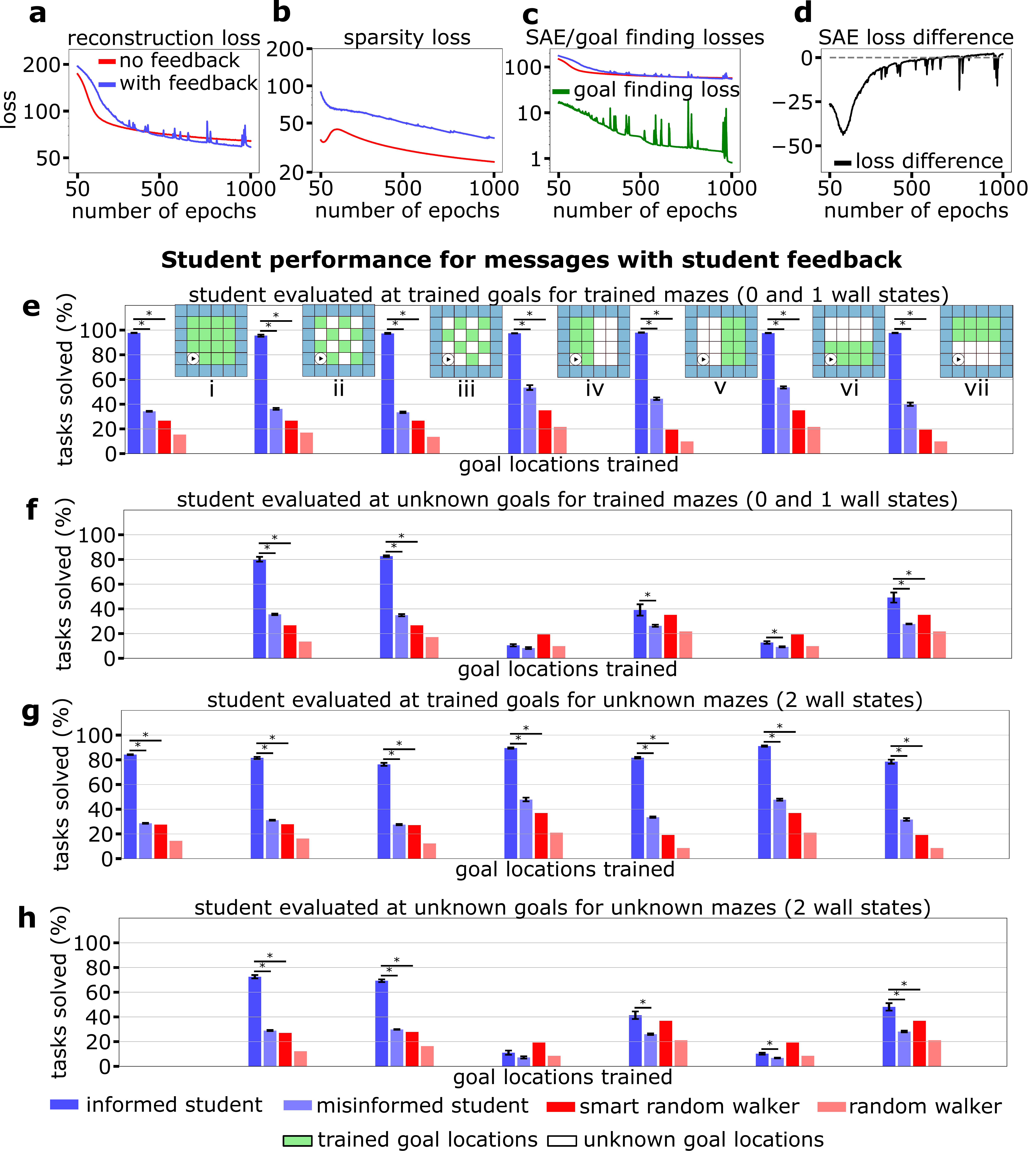}
    \caption{\textbf{Reward-based student feedback enhances performance, generalizability, and autoencoder reconstruction.} \ {\bf a)-d)} Components of the compound autoencoder training losses with and without student feedback; {\bf a)} reconstruction loss, {\bf b)} sparsity loss and {\bf c)} SAE loss, which additionally includes the goal finding loss when student feedback is included (see \ceqref{eq:autoencoder_loss} and \ceqref{eq:goal_finding_loss}). \emph{ d)} shows the difference $\mathcal{L}_{\textrm{SAE}}-\mathcal{L}_{\textrm{SAE, feedback}}$ which highlights that the student-feedback autoencoder achieves a lower reconstruction loss.
    {\bf e)-h)} Student performance on training and test maze tasks (see \hyperref[sec:methods]{Methods} for a description of the tasks). A comparison is made between the informed student, who receives the correct message, the misinformed student, who receives a message corresponding to a random task, and two random walkers, one of which never walks into walls (smart random walker). The performance is further evaluated for seven sets of trained goal locations, \emph{i)-vii)}, displayed as the green squares in the inset figures of {\bf e)}. In {\bf e)}, the performance is measured for the trained goal locations and trained mazes with 0 or 1 wall state. {\bf f)} shows the solve rate for the unknown goal locations (white) for mazes with 0 or 1 wall state. {\bf g)} and {\bf h)} depict the solve rates for new mazes (2 wall states) with trained and unknown goal locations, respectively. The error bars for each case refer to the variance across languages (five languages were trained for each case). }
    \label{fig:student performance and loss}
\end{figure}

In order to test the performance and generalization capabilities of the student, we used messages from teachers who mastered mazes with zero or one wall state and trained the student on patterned subsets of their goal locations (\figref{fig:student performance and loss}e, inset). We define the task solve rate as the percentage of goals attained under $2k_{\textrm{opt}}$ steps, where $k_{\textrm{opt}}$  corresponds to the shortest path from start to goal (see \hyperref[sec:methods]{Methods}). 

Under these terms, we can observe an increased performance of the student against misinformed students (given incorrect messages) and random walkers (one of which avoids walls) when evaluating the trained goal sets (\figref{fig:student performance and loss}e). We note that even in this scenario, this misinformed student slightly outperforms the random walkers, which we hypothesize is because of the initial action preference we observe for all messages, which allows the misinformed student to avoid the outer walls. To ascertain whether the generalization of the goal locations across the messages was achieved, we tested the performance of the student on unknown goals. We observe that the best generalization is achieved under checkerboard patterns. However, the performances of the other four cases do not differ significantly from the random walkers (\figref{fig:student performance and loss}f). This implies that generalization is difficult when large portions of the task are unknown, and interpolation between known states is not possible. Training on far-away goal locations leads to slightly better performance (\figref{fig:student performance and loss}f(v) and (vii)), but this might also be due to a wall-avoiding action preference. In this respect, when adding new wall locations, the overall performance is reduced, but the improvement against the other agents is preserved (\figref{fig:student performance and loss}g and h). These results highlight the importance of the goal-oriented structure of the lower-dimensional representations for these tasks and reinforce the benefits of the altered language achieved by the student feedback. In line with previous observations (\figref{fig:nonlinear message space}b), the main features of the encoded message are the policy and goal location. Therefore, when the agent attempts to solve the maze with unknown goals, it performs markedly worse than before. This behavior is only avoided when the student is trained on the checkerboard pattern, which means it has seen the entire maze and can use the information presented and its own experience to compensate for the lack of information. In other words, new tasks must be composable from other tasks within the language framework for communication to succeed.

\subsection*{Closing the loop}

As natural language is not usually restricted to sender and receiver but is a robust exchange between two agents, our final analysis is related to studying the effect of passing task information gained by the student through the language encoding to obtain a set of novel messages. Rather than solely relying on a set of teachers that perform single tasks and pass on compressed information, we allow the student to generate messages itself after performing – and thus learning – tasks with messages from teachers. These student messages are then passed back to itself, and its performance with these messages is assessed. A schematic depicting this structure can be seen in \figref{fig:model_setup}d. Thus, we attempt to create a simple generalist agent to supply information through the same language encoding, which we keep fixed. This communication process will naturally erode the message, leading to comparisons to the children's game ``telephone". In that setting, what information is robust to communication erosion is often studied. We can use this analogy also to identify the type of information that is more transmissible between agents \cite{mesoudi2008multiple}.

\begin{figure}[h!]
    \centering
    \includegraphics[width=.9\linewidth]{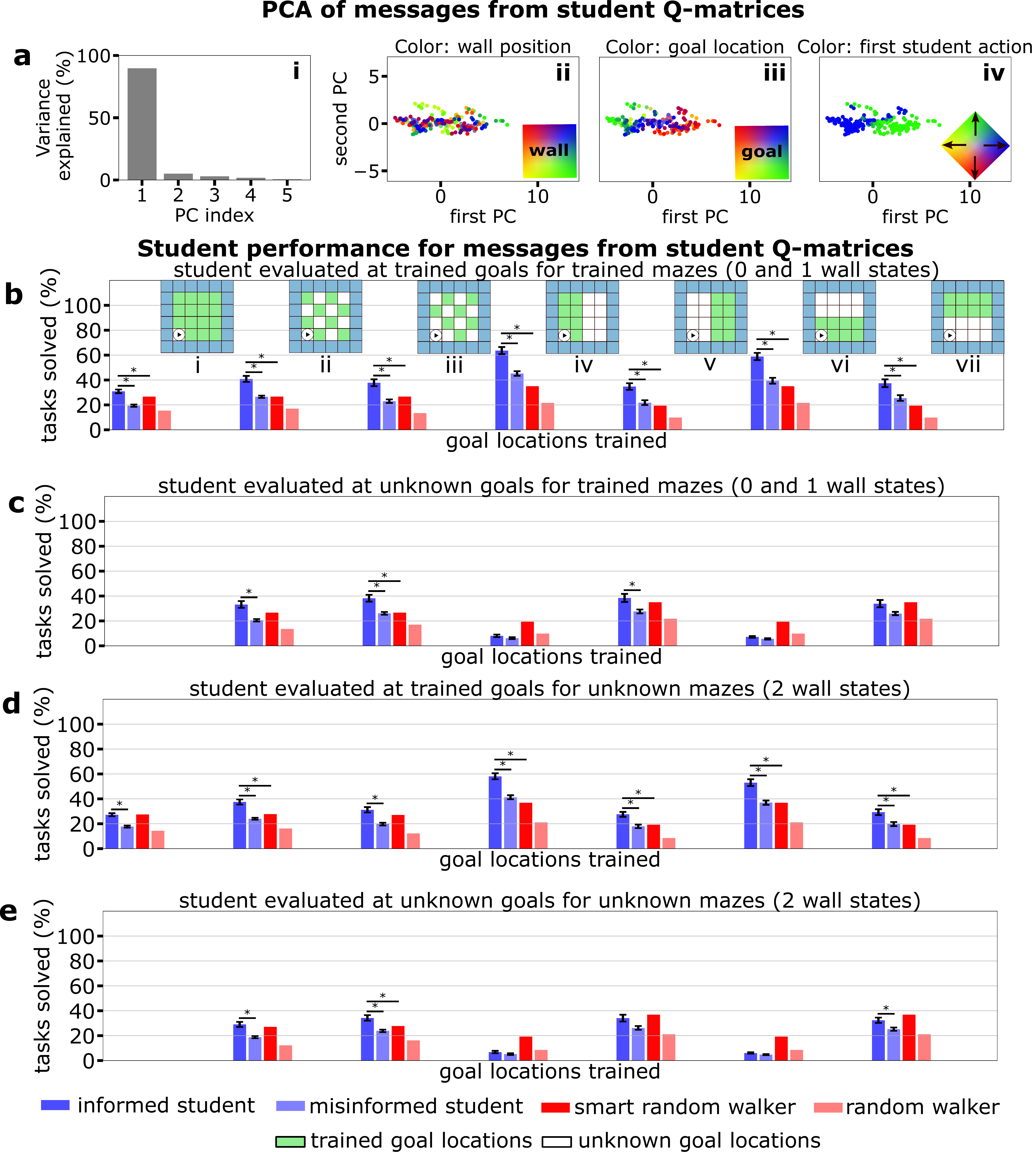}
    \caption{\textbf{Student-to-student communication leads to lower performance, but relevant task information is maintained.} \ {\bf a)} PCA on the \grqq{}degraded messages\grqq{} (encoded messages arising from the student task information): \emph{i)} shows the variance explained by PC. \emph{ii)-iv)} depict the degraded messages marked by wall position, goal location and probability of initial student action, respectively. {\bf b)-e)} Informed student performance on training and test maze tasks (see \hyperref[sec:methods]{Methods} for details) is compared against the misinformed student and two random walkers. The comparison is once more performed for the seven sets of trained goal locations, \emph{(i)-(vii)}. The relevant tasks per panel are identical to \figref{fig:student performance and loss}. For (b-e), 25 languages were originally trained and evaluated, but a subset was excluded (see \hyperref[sec:methods]{Methods} for details on this exclusion)
    }
    \label{fig:closing loop}
\end{figure}

Firstly, we can observe a degradation of the information content. Notably, the low-dimensional form of the student-generated task information entails that variability among student messages is mainly concentrated on a single dimension that is identifiable by the goal location and initial action (\figref{fig:closing loop}a(i), Table \ref{tab:variances_messagesstudentQ}). This contrasts with previous findings that the message space of teachers was not dominated by one principal component, and variability also corresponded to wall arrangements (\figref{fig:nonlinear message space}b). We then turn to the task completion rates of the students. Here both the informed and misinformed students are given messages resulting from encoding the task information the student has learned when supplied the teacher messages. The informed student is supplied with the encoded message corresponding to the current task, while the misinformed student is provided a message from a random task. From a performance perspective, we note that the degradation of the message content translates into lower task solve rates (\figref{fig:closing loop}b-e). This decrease can be seen even when considering trained goal locations (\figref{fig:closing loop}b). Nevertheless, students performed better than the misinformed agents, which implies that passed degraded messages include sufficient information to avoid walls and find the goal state. 

Given that the key features of the message arising from the lower-dimensional representations are the goal location and initial action features, it is unsurprising that, as long as the goals are known, the informed student performs well on maze tasks it has not seen before (\figref{fig:closing loop}d). When considering the performance of the student on unknown goals, for both trained (\figref{fig:closing loop}c) and untrained goal locations (\figref{fig:closing loop}e), we note that, in most cases, the informed student performs, at most, on par with the smart random walker. This indicates that the message has a detrimental effect on the students. It cannot generalize to the goals it has not seen, as the information provided does not allow it to build an adequate representation of the task. Finally, even though the degraded messages do not carry significant information about the world configuration, we hypothesize it is sufficient to produce minimally better performance in the known worlds compared to the unknown worlds.

We conclude that the student output retains pertinent task information that can enhance the performances of other students, even if degraded. This can be seen in the solve rates of the informed student. They are always higher than those of the misinformed student, allowing us to assume that the student can use relevant information within the degraded message. However, generalizability to unknown goals is lost under this framework, even when the student previously achieved high success rates (\figref{fig:student performance and loss}f and h, checkerboard).

Nevertheless, these results represent an early attempt to analyze task-driven communication with generalist agents. Particularly, one key aspect is how a compromise or balance between tutoring and learning can be achieved in multi-task and multi-agent systems to keep a relevant and generalizable message space. In other words, relevant features across tasks can be captured by a centralized embedding generated by individual experiences of agents, similar to how biological agents behave.

\begin{table}[h!]
    \centering
    \begin{tabular}{l c c c c}
        \toprule
        \textbf{Message grouping} & \textbf{$\textrm{Var}_{\textrm{within}}(X)$} & \textbf {$\textrm{Var}_{\textrm{between}}(X)$} & $\beta$ & F-value \\
        \midrule
        By wall position (\figref{fig:closing loop}a(ii)) & 1583 & 54.5 & 0.033 & 0.48\\
        By goal location (\figref{fig:closing loop}a(iii)) & 367 & 1270 & 0.776 & 48.15$^*$\\
        \bottomrule
    \end{tabular}
    \caption{Analysis of variance for world groups and goal groups in the message spaces from \figref{fig:closing loop}a. Statistical analysis is described in the \hyperref[sec:methods]{Methods} section. \,$^*$ refers to a significant difference in group means with significance level set at $p=0.05$.}
    \label{tab:variances_messagesstudentQ}
\end{table}

\section*{Discussion}
Task-relevant representations, either in the brain \cite{behrens2018cognitive,flesch2023continual}, as part of a linguistic system \cite{tomasello2009cultural,ten2017assessing} or in artificial agents \cite{bengio2013representation}, ought to be generalizable. However, it remains open how social agents can reconcile abstractions from their own experience with those acquired through communication. This work shows that a simple RL multi-agent model, which uses supervised teacher-to-student communication, can account for agent-wide variability from individual tasks that present differences in goal and state spaces. Notably, a low-dimensional representation of features in the state-action value function produces competent abstractions that permit other agents to learn goal and state spaces flexibly, even if they originate from model-free teachers. Additionally, we present a framework to analyze the nature of such communication protocols. Using this framework, we studied the lower-dimensional representations of the message space, both those purely teacher-driven and ones where a utility function is dependent on the performance of the student. We found the latter not only improved performance but also yielded a latent structure that prioritized variability along the goal space instead of the maze configuration, in contrast with the prominence of the state space in the solely teacher-based one. This suggests that reward-based constraints, which obey the return of another agent, can reproduce the task structure by prioritizing some modes while acquiring a similar – or superior – reconstruction error.

The motivation for this study is inspired by possible aspects that characterize natural language emergence. Thus, to summarize the results of this paper and their possible impact, we reiterate these with possible analogies arising from natural language. First, our language evolves according to a utility or gain function, not only with respect to comprehensibility or error minimization. In other words, it is not sufficient that information or concepts can be decoded in another agent, but the message space should also be advantageous to that agent. This can be compared to the evolution of natural language, where morphemes change according to motives, goals, and efficient representations of speakers and listeners \cite{seyfarth2014evolution}. Nonetheless, our language disregards trademarks of evolutionary linguistics \cite{mcmahon2012evolutionary}. For example, we lack a clear syntax or grammar, which can be modeled as a combination of actions and objects \cite{nowak1999evolution}, that provides a hierarchical space between its different modes and relates to behavioral variables like goal position and action. The introduction of dimensionality and sparsity constraints was motivated by natural language's evolution under anatomical and cognitive limitations, such as vocal tract size or memory capacity\cite{fitch2010evolution, christiansen2016now}. Hence, by allocating a predefined number of dimensions to our communication system, we replicate such properties and observe that these are organized into hierarchical task-relevant modes.

Instead of treating our system in a fashion akin to linguistics, we approach the communication problem as one of the top-down representations, in which the individual cognitive map should follow a higher-order one provided by representations from communication. In opposition to previous analysis of spatial language \cite{Spranger2016}, we did not assume a preexisting lexical or categorical model but, in contrast, have model-free agents whose policies need to be represented at a generalist level so they can be acquired at the individual one. This is similar to social species that have shown a cultural or experience-dependent complexity in their linguistic traits, like non-primate mammals such as bottlenose dolphins \cite{JANIK1998829} or naked-mole rats \cite{barker2021cultural}. In this sense, we presume that the neural representations and circuitry of the agents evolve and rewire to enable social learning \cite{dunbar1998social}. By doing so, we look at how the community scale influences the cognitive one – and vice-versa – instead of fixing communication or neuronal representations.

Additionally, we studied the importance and sensitivity of this language space by feeding back the space-action value maps of students in a similar manner to the telephone game \cite{mesoudi2008multiple}. The degraded information confirms the relationship between the quality of representations of agents and their performance and also points to the importance of a good sample space to construct language. Despite this degradation, we retained certain features that were important for task performance, a similar effect that has been observed in human speech\cite{breithaupt2018fact}. Overall, our results indicate that a generalist agent should be able to relate back to the language space in an invariant manner. Furthermore, specific social structures, such as teacher or student roles, may be critical to a robust language space and to moderate the information flow. 

Therefore, a promising research question revolves around achieving generalist agents in an RL framework, i.e., capable of being both teachers and students. In this case, an intermediate step would first consider students with separable sender and receiver units while also developing an experience-based policy. Such a system can permit us to study the level of confidence given to the received information in contrast with self-experience and how the improved representations may improve the language structure. Another interesting direction concerns how the social graph intervenes in the language construction process; the information flow and source reliability are significant for constructing the communication space if these generalist agents are considered. Finally, and due to the flexibility of the introduced framework, it would be attractive to use additional environments and tasks beyond grid-world mazes. This navigational model establishes a good relationship between the cognitive map, linguistic representation, and other decision-making frameworks.

In conclusion, we have introduced a novel and effective approach to studying language emergence using reinforcement agents and an encoding network. We believe that the avenues opened by this research are as compelling as they are varied. Furthermore, we can use this framework to generate hypotheses and test these in a manner that is analogous to human speech. This work combines machine learning and linguistics insights, an approach that has provided novel insights into the characteristics that drive language emergence.

\clearpage
\section*{Methods}\label{sec:methods}

\subsection*{Teacher agent Q-learning}
In our communication model, shown in \figref{fig:model_setup}, the navigation task solutions are learned by the teacher agent (implemented via a multilayer perceptron) via deep Q-learning \cite{sutton2018reinforcement}. The Q-value for action $a$ and state $s$, $Q(s, a)$, represents the agent's future maximum return achievable by any policy. Despite the small state-action space, using an artificial neural network provides the flexibility to apply the framework to future tasks that may have much larger state-action spaces. Concretely, the teacher agents are trained to output Q-values satisfying the Bellman equation:
\begin{equation}\label{eq:bellman_equation}
	Q(s,a)=R^{s,a}+\gamma_{\textrm{Bellman}}\max_{a'}Q(s',a').
\end{equation}	   
Thus the expected future reward is composed of the immediate reward, $\mathcal{R}^{s,a}$, of the action, $a$, and the maximum reward the agent can expect from the next state $s'$ onward when behaving optimally, i.e., picking the action that promises the most reward. The temporal discount $\gamma_{\textrm{Bellman}}\in\left[0,1\right]$ signifies the uncertainty about rewards obtained for future actions ($\gamma_{\textrm{Bellman}}=0$ would be maximum uncertainty, we use $\gamma_{\textrm{Bellman}}=0.99$, see table \ref{tab:hyperparams}).

To train the DQN, we minimize Mean Squared Error (MSE) loss between the left- and right-hand sides of \ceqref{eq:bellman_equation}, i.e., we minimize
\begin{equation}\label{eq:q_network_learning}
	\mathcal{L}_{\textrm{DQN}}=\frac{1}{|\mathcal{T}|}\sum_{\langle s,a,R^{s,a}\rangle\in\mathcal{T}}|Q(s,a)-(R^{s,a}+\gamma_{\textrm{Bellman}}\max_{a'}Q(s',a'))|^2,
\end{equation}
where $\mathcal{T}$ is a set of transitions (state, action, and corresponding reward) $\langle s,a,R^{s,a}\rangle$ that the teacher DQN is trained on in the current optimization step. Thus, one optimization step is performed after each step the agent takes in the maze. The transition set $\mathcal{T}$ is composed of two distinct transitions: \emph{i)} \grqq{}long-term memory\grqq{} transitions, which are all unique transitions the agent has seen since training began, and \emph{ii)}  additionally weighted \grqq{}short-term memory\grqq{} transitions, which are the last $L$ transitions the agent has seen. Therefore, the transitions that have recently been executed several times have a higher impact on the loss function $\mathcal{L}_{\textrm{DQN}}$ than the ones that were encountered a long time ago.

\subsection*{Network specifications}
The student and teacher networks are identical multilayer perceptrons apart from the input dimension. Each neuron in the two networks (except for the $K$ message neurons) has a ReLU activation function and a bias parameter. The number of parameters per layer for the student and teachers is listed in table \ref{tab:studentteacher_params}.
\begin{table}
    \centering
    \begin{tabular}{l c c c c}
        \toprule
        \textbf{layer} & \textbf{neuron number} & \textbf{weight parameters} & \textbf{bias parameters} & \textbf{total parameters} \\
        \midrule
        input layer & $2+K'$ & - & - & -\\
        \midrule
        linear layer & $10$ & $20+10 K'$ & $10$ & $30+10 K'$ \\
        \midrule
        linear layer & $20$ & $200$ & $20$ & $220$ \\
        \midrule
        linear layer & $20$ & $400$ & $20$ & $420$ \\
        \midrule
        output layer & $4$ & $80$ & $4$ & $84$ \\
        \midrule
        \midrule
        total ($K'=0$) & $56$ & $700$ & $54$ & $754$ \\
        total ($K'=5$) & $61$ & $750$ & $54$ & $804$ \\
        \bottomrule
    \end{tabular}
    \caption{Network architecture in the teacher and student networks used in our toy model - in this context $K'$ is the number of extra network inputs in addition to the state's x- and y-coordinates. This $K'$ corresponds to the length of the message, i.e., $K'=K=5$ for the student, while the teacher does not receive a message, so $K'=0$.}
    \label{tab:studentteacher_params}
\end{table}
\newpage
The autoencoder neural network, which we use as a language proxy, consists of convolutional layers in addition to the fully connected layers. We use convolutions because the entries of the Q-matrix represent the states of the two-dimensional grid-world and, therefore, include spatial information that the network needs to learn. Thus, the input (Q-matrix of the teacher) is processed by two convolutional layers in the first half of the autoencoder, followed by one fully connected linear layer that outputs the message vector. After this dimensionality reduction, the decoding half of the autoencoder aims at reconstructing the original Q-matrix from the message vector. The architecture of the autoencoder is summarized in table \ref{tab:autoencoder_params}.
\begin{table}
\small
    \centering
    \begin{tabular}{l c c c c}
        \toprule
        \textbf{layer} & \textbf{neuron number} & \textbf{weight parameters} & \textbf{bias parameters} & \textbf{total parameters} \\
        \midrule
        input layer & $4\tilde{n}^2$ & - & $4\tilde{n}^2$ & $4\tilde{n}^2$ \\
        \midrule
        conv. layer & 10 filters (size $2\times2\times4$)  & 160 & 10 & 170 \\
        \midrule
        conv. layer & 10 filters (size $2\times2\times10$) & 400 & 10 & 410 \\
        \midrule
        linear layer & $K$ & $10(\tilde{n}+2)^2 K$ & $K$ & $10(\tilde{n}+2)^2 K+K$ \\
        \midrule
        linear layer & $10(\tilde{n}+2)^2$ & $10(\tilde{n}+2)^2 K$ & $10(\tilde{n}+2)^2$ & $10(\tilde{n}+2)^2(K+1)$ \\
        \midrule
        deconv. layer & 10 filters (size $2\times2\times10$) & 400 & 10 & 410 \\
        \midrule
        deconv. layer & 4 filters (size $2\times2\times10$) & 160 & 4 & 164 \\
        \midrule
        output layer & $4\tilde{n}^2$ & - & $\tilde{n}^24$ & $\tilde{n}^24$ \\
        \midrule
        \midrule
        total ($\tilde{n}=4$ and $K=5$) & 493 and 34 filters & 4720 & 527 & 5247 \\
        \bottomrule
    \end{tabular}
    \caption{Network architecture in the autoencoder network used in our toy model. In this context, $\tilde{n}$ is the maze dimensionality (we use $4 \times 4$ mazes, therefore, $\tilde{n}=4$) and $K$ is the length of the message (we use $K=5$).}
    \label{tab:autoencoder_params}
\end{table}

\subsection*{Training and test tasks}
\begin{figure}
    \centering
    \includegraphics[width=.6\linewidth]{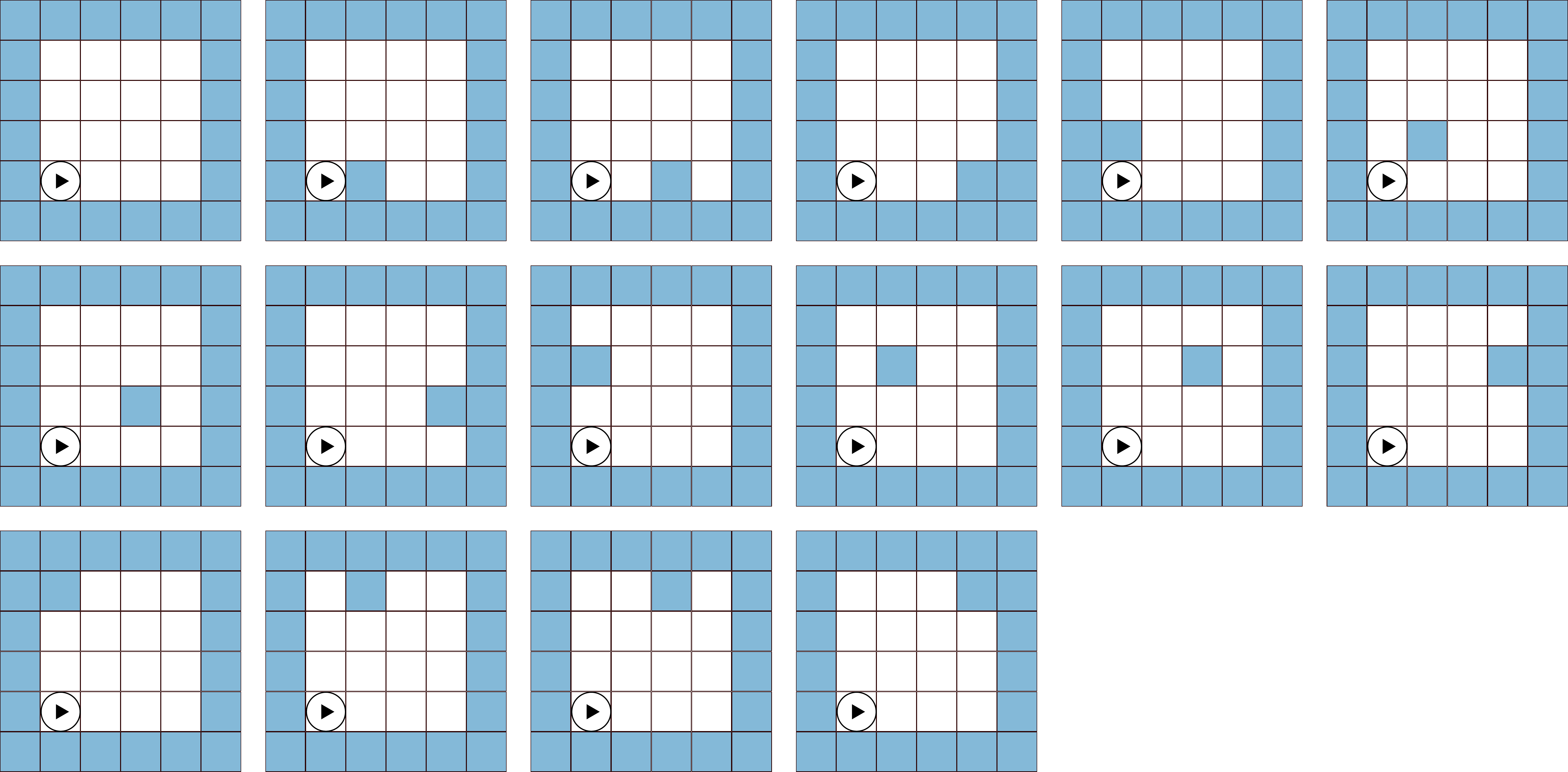}
    \caption{\textbf{All training mazes.} The student always starts in the bottom left corner. Light blue squares mark wall locations, which can not be accessed. The $4 \times 4$ mazes with 0 or 1 wall state comprise the training tasks.}
    \label{fig:all training mazes}
\end{figure}

\begin{figure}[h!]
    \centering
    \includegraphics[width=.6\linewidth]{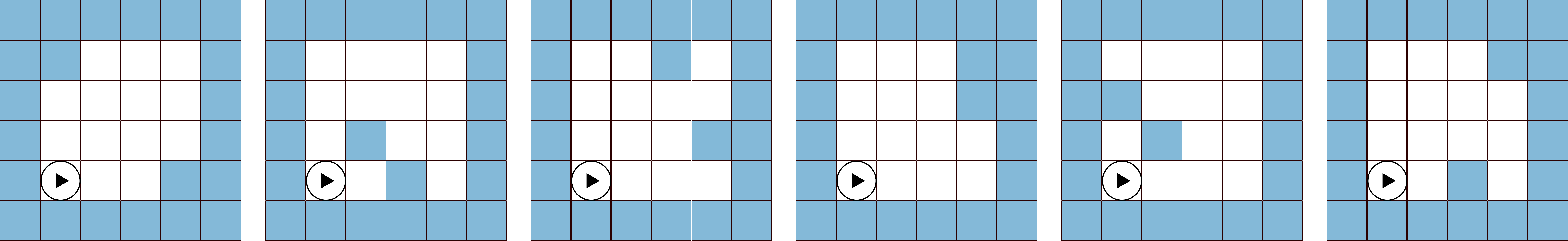}
    \caption{\textbf{Example test mazes.} The student always starts in the bottom left corner. Light blue squares mark wall locations, which can not be accessed. The $4 \times 4$ mazes with 2 wall states, except those where permissible states are cut off by walls, comprise the test tasks.}
    \label{fig:some test mazes}
\end{figure}

The square grid-world setting consists of a grid of size $n \times n$ (see examples in Figs. \ref{fig:all training mazes}, \ref{fig:some test mazes}). Given that each maze is surrounded by impenetrable walls, this gives the agent an effective number of possible states (including the initial state where the agent starts, the goal state the agent has to reach, and the wall states the agent can not cross) equal to $\tilde{n} \times \tilde{n}$, where $\tilde{n} = n-2$. In all cases, the agent starts in the bottom left corner. During the training of the SAE and student, we only include mazes with zero and one interior wall state, which gives us $( \tilde{n}^2 - 1)+(\tilde{n}^2 - 1)(\tilde{n}^2 - 2) =(\tilde{n}^2 - 1)^2$ possible maze-solving tasks. The agent moves through the grid-worlds with four discrete actions: single steps to the right, up, left, and down. Each episode starts with the agent at the initial state and ends when the goal is reached or the maximum number of steps has been taken. To avoid potential infinite loops or movements into the walls, the agent receives a small negative reward for any action ($R_{\textrm{step}}=-0.1$) and a large negative reward for hitting any wall ($R_{\textrm{wall}}=-0.5$). If the agent reaches the goal, they receive a large positive reward ($R_{\textrm{goal}}=2$). 

We used all $4 \times 4$ grid-worlds with 0 or 1 wall state, amounting to 16 worlds in total, see \figref{fig:all training mazes} as tasks for training the language and the student. In world 0 (top left), there are 15 possible tasks, i.e., goal locations, namely all states that are neither a wall nor the starting location (all white squares without inset in the figure). Similarly, in the 15 worlds with a single wall, there are 14 possible tasks, amounting to 225 tasks used for training the language and the student agent. During the teacher training, the Q-values of the wall state positions of the teacher Q-matrix are set to 0, as the agent can never visit them (due to bounce back).

For unknown tasks, we chose all possible configurations of mazes with two wall states, six examples of which are shown in \figref{fig:some test mazes}. We eliminated mazes that led to inaccessible states, leading to 101 possible configurations with two walls, each with 13 goal locations. Therefore, the test set was made up of 1313 test tasks in total.

\subsection*{The full autoencoder loss}
The loss function for the SAE (which does not include student feedback) is defined as:
\begin{equation}\label{eq:autoencoder_loss}
    \mathcal{L}_{\textrm{SAE}}=\mathcal{L}_{\textrm{reconstruction}}+\mathcal{L}_{\textrm{sparsity}}=(1-\kappa)||\tilde{Q}-Q||_{2}+\kappa||m||_{2},
\end{equation}
where $\kappa$ is a hyperparameter, which we can adjust to increase the importance of either the reconstruction or sparsity, $Q$ is the input Q-matrix, $Q'$ the reconstruction of the autoencoder, and $m$ is the lower-dimensional message.

We also included the student in the training of the language. Therefore, we augmented the autoencoder loss to include a term that enforced the usefulness of the messages to the student. This was done by first generating the student output for each possible state $s=(x_{s},y_{s})$, which consisted of four real numbers representing the four possible actions. Applying a softmax function to those values, we obtained action probabilities for the four actions in each state. Given all the action probabilities, we could calculate the state occupancy probabilities for the student after any number of steps $k$. We then defined the solve rate of a task as the state occupancy probability of the goal state after $k$ steps, as this state could not be left once it had been reached. We aimed for optimal solutions to be found; therefore, we always allowed the student only $k=k_{\textrm{opt}}$ steps to solve the task during training, where $k_{\textrm{opt}}$ was the length of the shortest path to the goal.

This amounts to the first term in \ceqref{eq:goal_finding_loss} of the student goal finding loss. The exponent was chosen to avoid the local minimum of the loss in which a small number of training tasks are not solved at all while the majority is solved perfectly. The second term in \ceqref{eq:goal_finding_loss} is a regularization of the student output while the hyperparameter $\gamma$ controls the relation between the two parts.
The regularization of the student Q-matrix is also normalized by the number of its entries $4\tilde{n}^2$. 

\begin{equation}\label{eq:goal_finding_loss}
    \mathcal{L}_{\textrm{goal finding}}=(1-\gamma)\left(1-\mathbb{P}\left[s_{k}=s_{\textrm{goal}}\right]\right)^4+\gamma\frac{||Q_{\textrm{student}}||_{2}}{\sqrt{4\tilde{n}^2}}
\end{equation}

\subsection*{Analysis of variance in the message spaces}
We analyze the structure of the different message spaces by studying the relative variances explained by the two features describing each navigation task: the placement of the walls and the goal's location. 

In this context, two types of variance can be computed: a \textit{variance within groups} and a \textit{variance between groups}, where a group is made up of either all tasks within a maze (i.e. same wall position) or all tasks with the same goal location. The former variance is lower when each group is clustered tightly, but the distance between groups is large. The latter variance is lower when the means of the groups cluster tightly, but there is a larger data spread within each group. To simplify the equations that follow, we introduce $M$, the total number of messages, $N$, the number of distinct groups, $M_i$, the number of elements in group $i$ and $m_{ij}$, which refers to the $j$-th message of group $i$. Then, the mean of each group is $\bar{m}_{i}=\frac{1}{M_{i}}\sum_{j=1}^{M_{i}}m_{ij}$ and the overall mean is $\bar{x}=\frac{1}{M}\sum_{i=1}^{N}\sum_{j=1}^{M_{i}}m_{ij}$. Thus the variance within and between groups of messages is defined by
\begin{align}
    \textrm{Var}_{\textrm{within}}(X)&=\sum_{i=1}^{N}\sum_{j=1}^{M_{i}} (m_{ij}-\bar{m}_{i})^{2}
    \label{eq:within_group_variance} \\
    \textrm{Var}_{\textrm{between}}(X)&=\sum_{i=1}^{N}M_{i}(\bar{m}_{i}-\bar{m})^{2}
    \label{eq:between_group_variance}
    \\
    \beta &=\frac{\textrm{Var}_{\textrm{between}}}{\textrm{Var}_{\textrm{within}}+\textrm{Var}_{\textrm{between}}}
    \label{eq:beta_variance}
\end{align}
Here, we introduce a value $\beta$, which allows for a comparison between the two different variances. When $\beta$ is close to 1, the variance between groups dominates and vice versa. 

Using the above variances, we can statistically test whether the means of all message groups (grouping either by wall position or goal location) are significantly different from each other by introducing the concept of the F-value, which is defined as the ratio of the mean square distance between groups $MS_{B}$ and the mean square distance within groups $MS_{W}$:

\begin{align}
    MS_{B}&=\frac{\textrm{Var}_{\textrm{between}}}{N-1}, \label{eq:meansquare_between} \\
    MS_{W}&=\frac{\textrm{Var}_{\textrm{within}}}{M-N}, \label{eq:meansquare_within}\\
    F&=\frac{MS_{B}}{MS_{W}}. \label{eq:fvalue}
\end{align}
The group means differ significantly when the F-value is greater than a critical F-statistic (depending on a significance threshold $p$ and the degrees of freedom). As we removed the world with no wall states in the analysis of variances, the values of  $F_{\textrm{crit}}$ (listed in table \ref{tab:critical_fvalues}) are the same in both grouping cases (by maze and by goal). The two degrees of freedom are $N-1=14$ and $M-N=195$.

\begin{table}[h!]
    \centering
    \begin{tabular}{c c}
        \toprule
        \textbf{Significance threshold $p$} & \textbf{critical F-value $F_{\textrm{crit}}$} \\
        \midrule
        0.1 & 1.54 \\
        0.05 & 1.74 \\
        0.01 & 2.17 \\
        0.005 & 2.35 \\
        0.001 & 2.74 \\
        \bottomrule
    \end{tabular}
    \caption{Critical F-values for our data groupings by wall and goal for different significance thresholds.}
    \label{tab:critical_fvalues}
\end{table}

\subsection*{Statistical methods}
One-sample t-tests were used when comparing the informed and misinformed students against the smart random walker, and a  two-sample t-tests when comparing against the misinformed student.

\subsection*{Language selection}
Initially, 25 languages were trained and evaluated when the student information was encoded and used for the navigation maze in \figref{fig:closing loop}. However, within that set of languages, we encountered a subset of languages (approximately 30\%) that led to lower solving rates for the informed student than the misinformed student or random walker on the trained tasks. The structure of these languages, which we defined as inefficient, led to a loss of task-critical information during the encoding. These languages were removed from the set of languages we analyzed in \figref{fig:closing loop}. We argue that this is akin to the effect of natural evolution, where weak and inefficient members (in this case, languages) do not survive. Therefore, our criteria for the "survival of the fittest" language is the following: if the language leads to an average task-solving rate for the informed student (receiving the message from encoded student information) higher than the average solving rate of the misinformed student and the random walker (all measured on the trained tasks), it survives.

\section*{Acknowledgements}

We acknowledge the support of the {\bf Institute of Experimental Epileptology and Cognition Research} at the 
University of Bonn Medical Center and {\bf Institute for Physiological Chemistry} at the University of Mainz Medical Center and {\bf Joachim Herz Foundation}. We thank {\bf Alison Barker} and {\bf Martin Fuhrmann}  for fruitful discussions, and all members of the Tchumatchenko group, particularly {\bf Pietro Verzelli}, for feedback on the manuscript.

\section*{Data and code availability}
Computer code to train the agents, generate languages and plot the figures can be found in the following public github repository www.github.com/meggl23/multi\_agent\_language, \emph{DOI:10.5281/zenodo.7885527}.

\bibliography{main}

\setcounter{figure}{0}
\renewcommand{\thefigure}{S\arabic{figure}}
\setcounter{table}{0}
\renewcommand{\thetable}{S\arabic{table}}
\newpage
\section*{Supplemental material}
\subsection*{The effect of linearity in the autoencoder and the student}
\begin{figure}[h!]
    \centering
    \includegraphics[width=.9\linewidth]{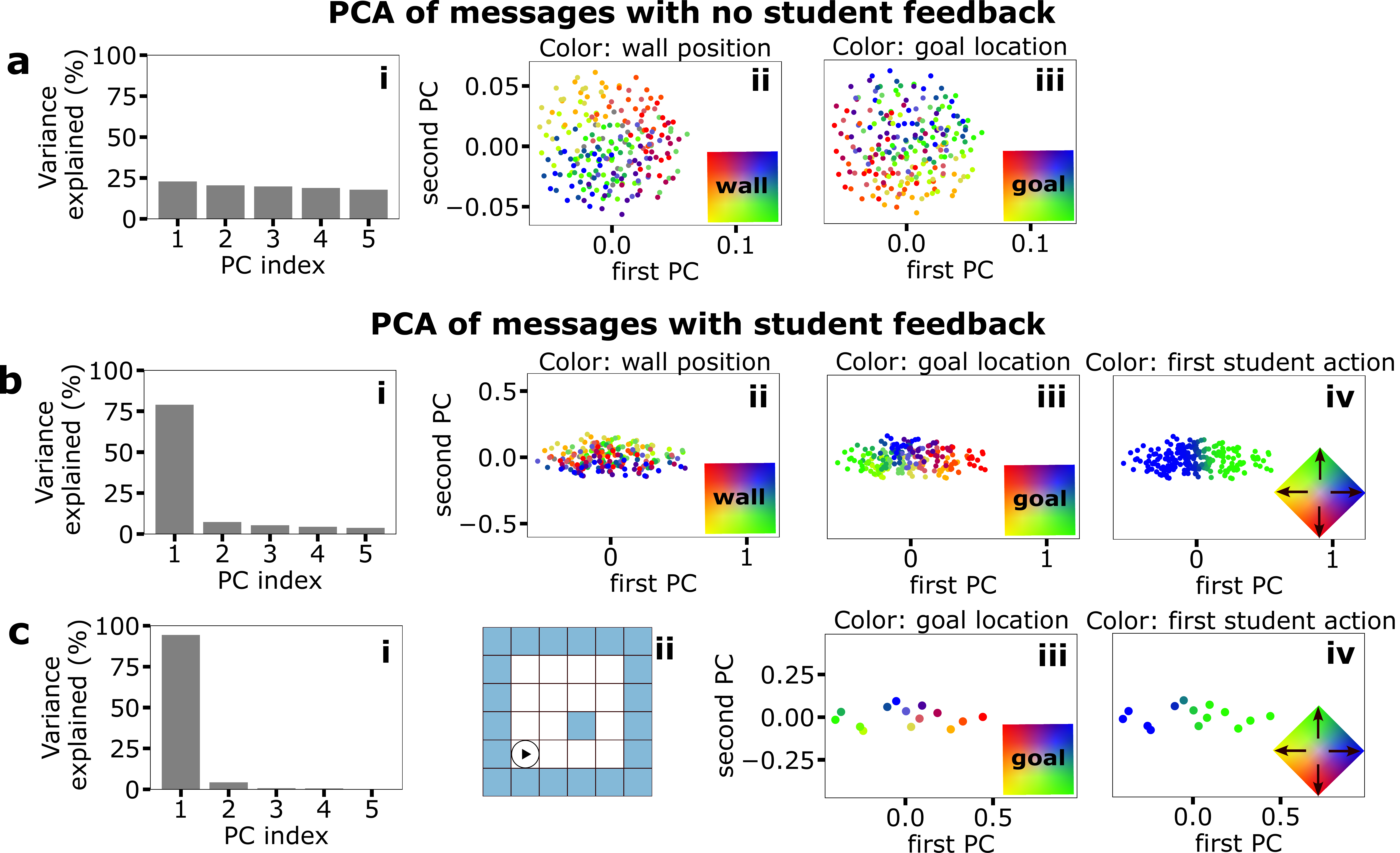}
    \caption{\textbf{Removing the non-linearities of the autoencoder and student networks leads to a significantly altered embedding structure.} \ {\bf a)} 
    Principal Component Analysis (PCA) of the lower-dimensional messages of size ($K=5$) obtained from a language encoding without student feedback (\ceqref{eq:autoencoder_loss}) for all possible tasks in the $4 \times 4$ mazes with $\leq1$ walls. {\it i)} Explained variance by principal component. {\it ii)-iii)} depicts the messages highlighted by the position of the single wall (gray refers to the maze with no walls) and by the position of the goal, respectively. {\bf b)} Result of the message encoding now including student feedback achieved by using \ceqref{eq:student_loss} for the loss function. {\it i)-iii)} depict the same concepts as in \emph{a)}. {\it iv)} shows messages highlighted by preferred first student action (step up or right). {\bf c)} PCA of the messages with student feedback from an example grid-world (depicted in {\it (ii)}).}
    \label{fig:all linear message space}
\end{figure}

\begin{figure}
    \centering
    \includegraphics[width=.9\linewidth]{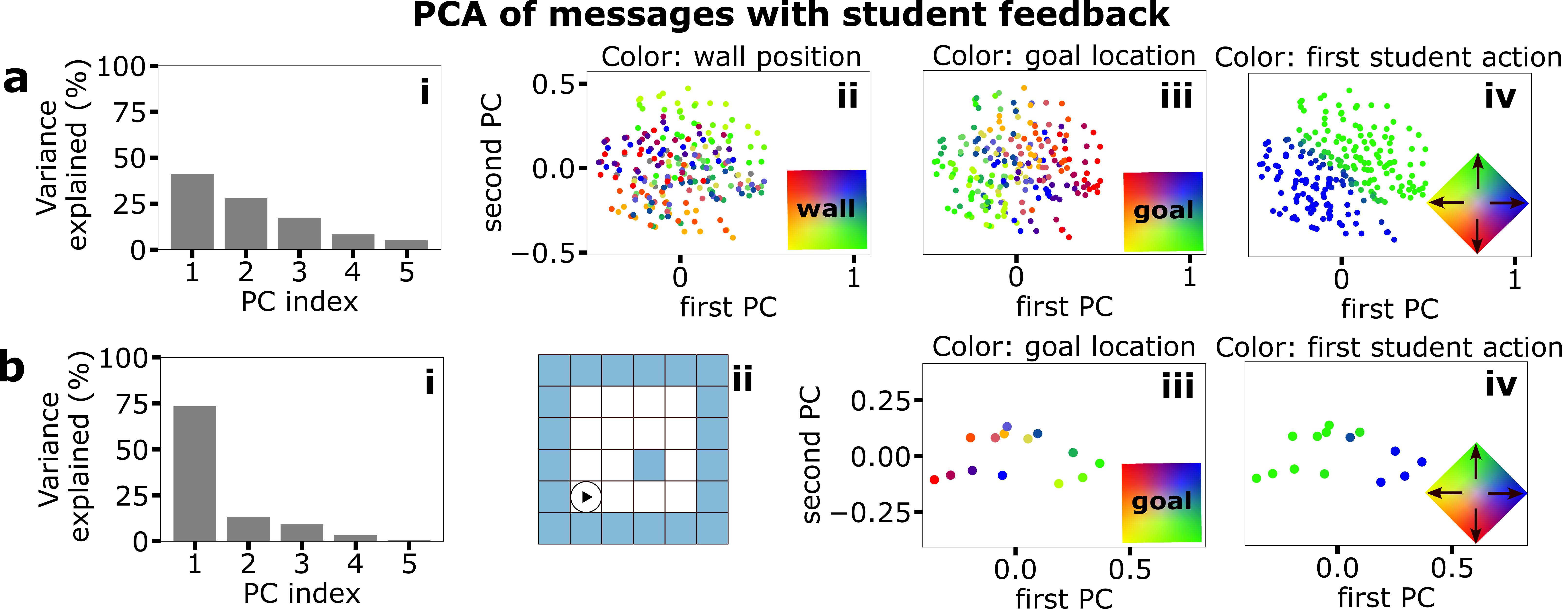}
    \caption{\textbf{Reintroducing the non-linearity in the student and then applying student feedback, leads to an approximation of the structure of the full non-linear setup.} \ {\bf a)} 
    Principal Component Analysis (PCA) of the lower-dimensional messages of size ($K=5$) obtained from a language encoding with student feedback (\ceqref{eq:autoencoder_loss}) for all possible tasks in the $4 \times 4$ mazes with $\leq1$ walls. {\it i)} Explained variance by principal component. {\it ii)-iv)} depicts the messages highlighted by the position of the single wall (gray refers to the maze with no walls), by the position of the goal, and by preferred first student action (step up or right). {\bf b)} PCA of the messages with student feedback from an example grid-world (depicted in {\it (ii)} ).}
    \label{fig:autoenc linear message space}
\end{figure}

All our networks in our results were implemented with non-linear activation functions, so for completeness' sake we include the results arising from removing those non-linearities. The results of having linear student and autoencoder architectures can be seen in \figref{fig:all linear message space}, while a non-linear student and linear autoencoder is shown in \figref{fig:autoenc linear message space}.

\clearpage

\subsection*{Student trained on \grqq{}frozen\grqq{} messages}

\begin{figure}[h!]
    \centering
    \includegraphics[width=.84\linewidth]{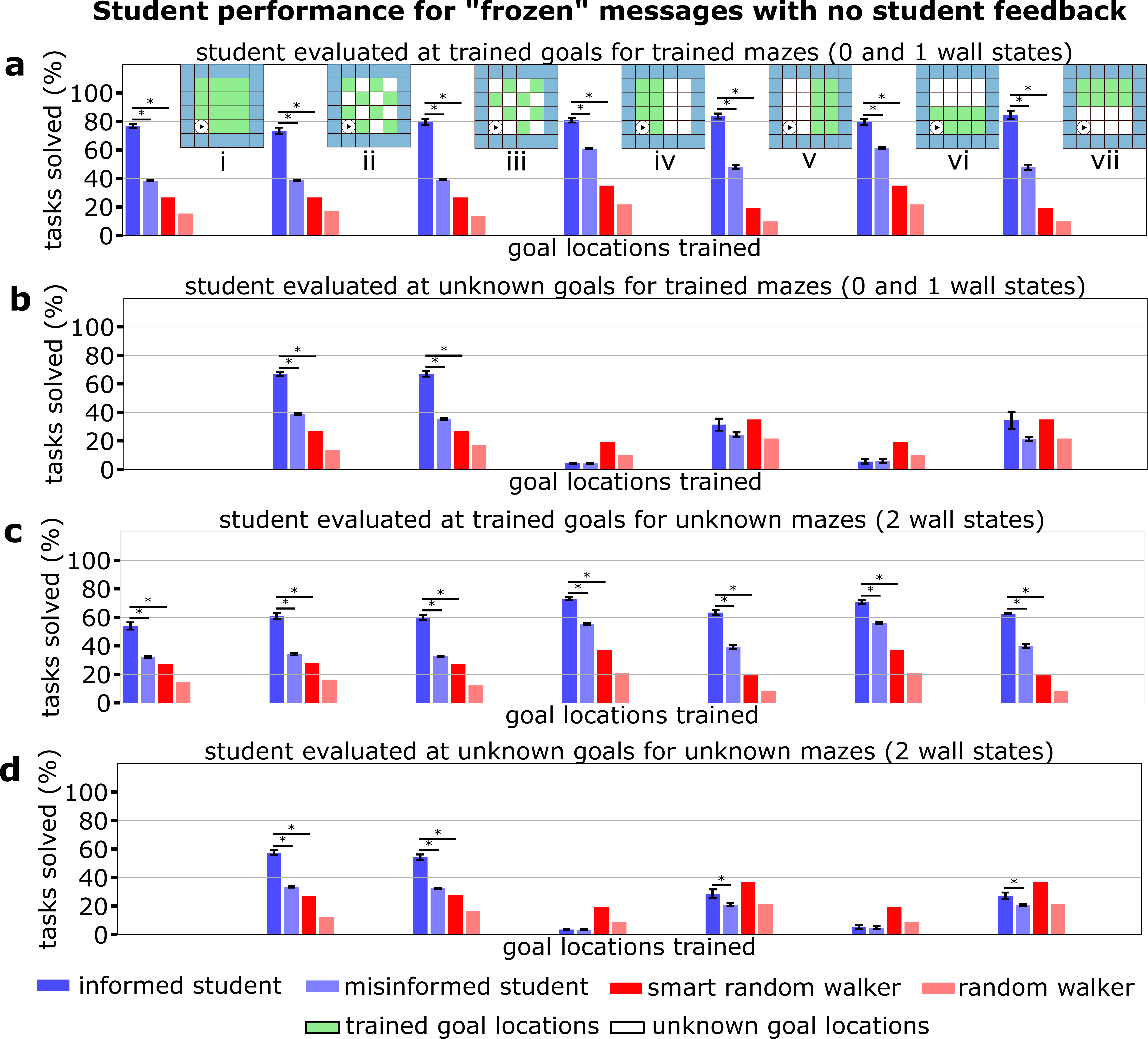}
    \caption{\textbf{Training a student on a set of frozen lower-dimensional representations, arising from an independently trained language (without feedback), leads to reasonable task performance. However, this performance is worse than the student-feedback version (cf. \figref{fig:student performance and loss}).} 
    {\bf a)-d)} A student is trained on a set of frozen messages arising from an independently trained language (created without feedback) and then compared against a misinformed student and two random walkers. The comparison is once more performed for the seven sets of trained goal locations, {\it (i)-(vii)}. The relevant tasks per panel are identical to \figref{fig:student performance and loss}.}
    \label{fig:student performance frozen1}
\end{figure}

\begin{figure}[h!]
    \centering
    \includegraphics[width=.84\linewidth]{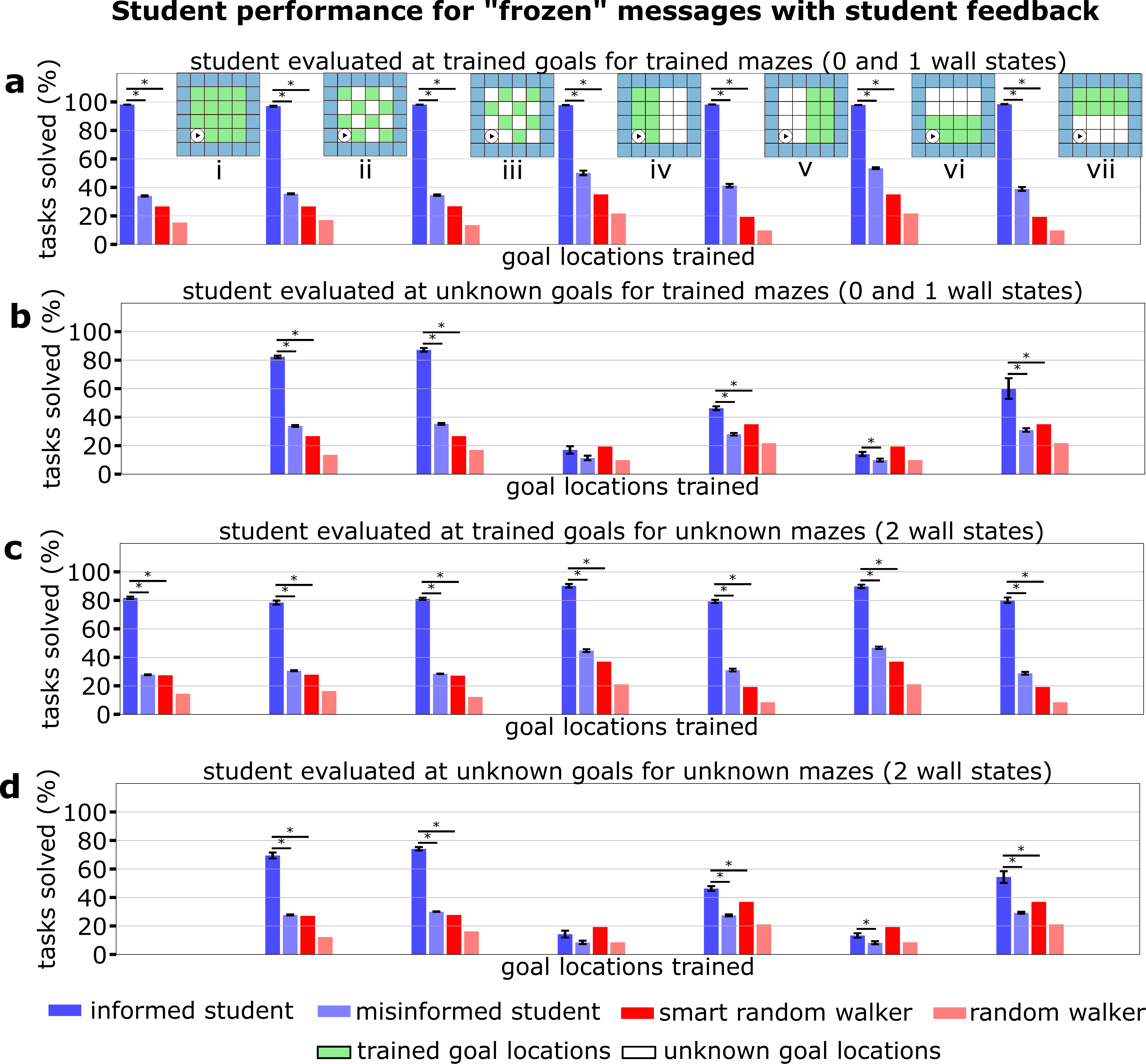}
    \caption{\textbf{Training a student on set of frozen lower-dimensional representations arising from an independently trained language (with feedback), leads to better performance than when the language is allowed to change mid-training (cf. \figref{fig:student performance and loss}).} 
    {\bf a)-d)} A student is trained on a set of frozen messages arising from an independently trained language (created with feedback) and then compared against a misinformed student and two random walkers. The comparison is once more performed for the seven sets of trained goal locations, {\it (i)-(vii)}. The relevant tasks per panel are identical to \figref{fig:student performance and loss}.}
    \label{fig:student performance frozen2}
\end{figure}

\subsection*{Autoencoder loss plots for different hyperparameters}
\begin{figure}[h!]
    \centering
    \includegraphics[width=.95\linewidth]{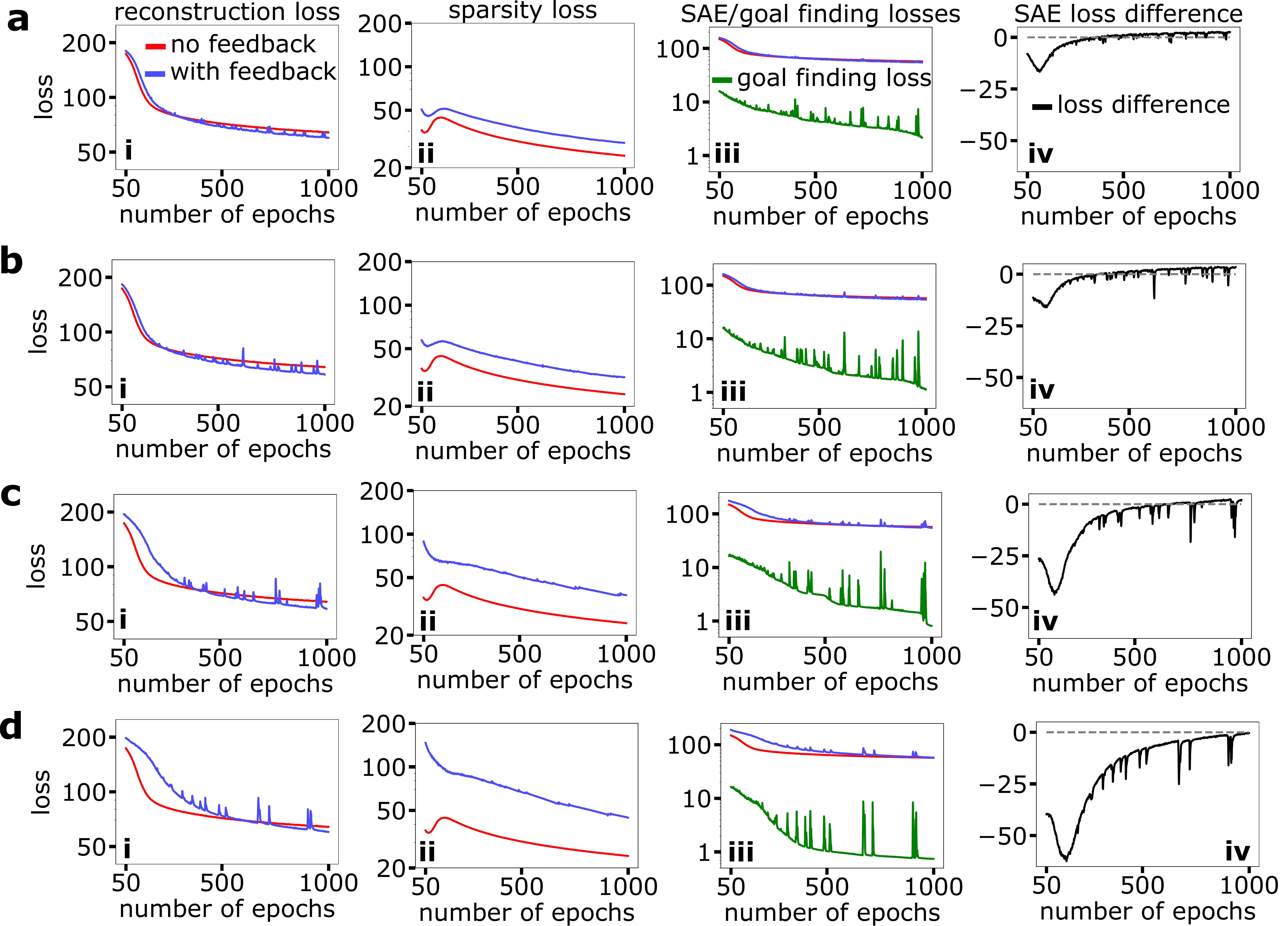}
    \caption{\textbf{The compound autoencoder with student feedback loss is lower for a range of different values of the hyperparameter $\zeta$ (the student-feedback weighting)}. Each row ({\bf a)-d)}) represents a different value for $\zeta=1,2,5,10$. From left to right the plots show (i) $\mathcal{L}_{\textrm{reconstruction}}$ and $\mathcal{L}_{\textrm{reconstruction, feedback}}$, (ii) $\mathcal{L}_{\textrm{sparsity}}$ and $\mathcal{L}_{\textrm{sparsity, feedback}}$, (iii) $\mathcal{L}_{\textrm{SAE}}$, $\mathcal{L}_{\textrm{SAE, feedback}}$ and $\mathcal{L}_{\textrm{goal finding}}$, (iv) $\mathcal{L}_{\textrm{SAE}}-\mathcal{L}_{\textrm{SAE, feedback}}$. For reference see  \ceqref{eq:student_loss} and \ceqref{eq:autoencoder_loss} .}
    \label{fig:autoenc loss plots}
\end{figure}

The robustness of the results from \figref{fig:student performance and loss}a-d was checked by varying the hyperparameter $\zeta$, which controls the relative importance of the autoencoder (SAE) and student goal finding losses (\ceqref{eq:student_loss}). The results in the main text (\figref{fig:student performance and loss}a-d) were obtained with $\zeta=5$, but \figref{fig:autoenc loss plots}a(iv)-d(iv) show that the results hold for $\zeta\in\{1,2,10\}$ as well.

\clearpage
\subsection*{Teacher and student Q-matrices}

We also performed PCA directly on the teacher and student Q-matrices, to see if the structures and features that emerged were comparable to the language encoding.

\begin{figure}[h!]
    \centering
    \includegraphics[width=.95\linewidth]{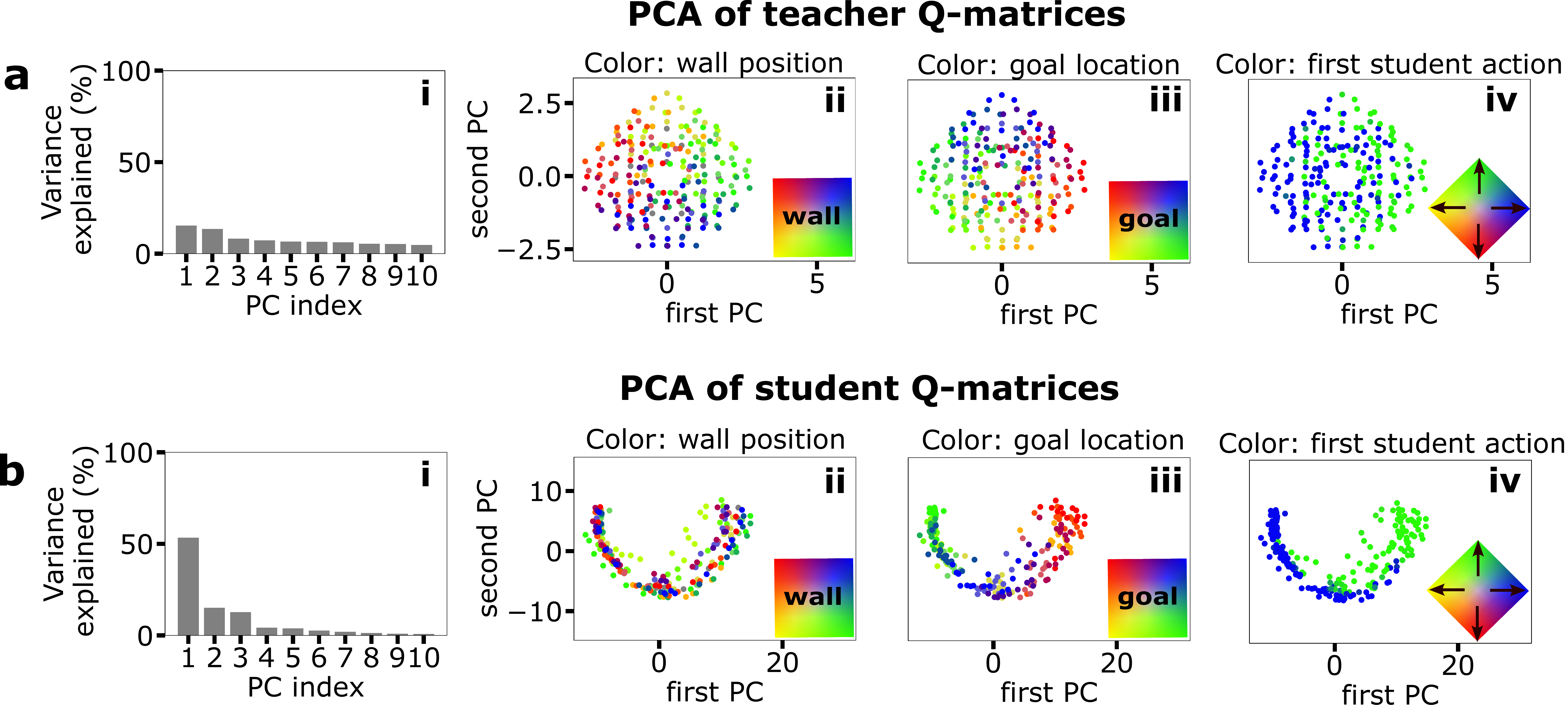}
    \caption{\textbf{PCA applied directly to the task information from the student and teachers reveals remearkable structure that mirrors the tasks themselves.} {\bf a)} PCA of the Q-matrices learned by the teacher networks. Each data point corresponds to a Q-matrix and, therefore, to a maze task. From left to right the figures show {\it (i)} the variance explained by the first 10 PCs, and projections of the matrices to the first two PCs with coloration by {\it (ii)} maze identity, {\it (iii)} goal location and {\it (iv)} initial student action. {\bf b)} PCA of the student \grqq{}Q-matrices\grqq{}, which are task information matrices learned by the student, but correspond to action probabilities instead of correct Q-values. The subfigures are identical to (a). The data dimensionality in both cases is $\tilde{n}\times \tilde{n}\times 4$ ($=64$ for maze size $\tilde{n}=4$). }
    \label{fig:teacher student Q}
\end{figure}

\begin{table}[h!]
    \centering
    \begin{tabular}{l c c c c}
        \toprule
        \textbf{Message grouping} & \textbf{$\textrm{Var}_{\textrm{within}}(X)$} & \textbf {$\textrm{Var}_{\textrm{between}}(X)$} & $\beta$ & F-value \\
        \midrule
        By wall position (\figref{fig:teacher student Q}a(ii)) & 552 & 1895 & 0.774 & 47.81\\
        By goal location (\figref{fig:teacher student Q}a(iii)) & 1829 & 618 & 0.253 & 4.71\\
        By wall position (\figref{fig:teacher student Q}b(ii)) & 23235 & 5147 & 0.181 & 3.09 \\
        By goal location (\figref{fig:teacher student Q}b(iii)) & 9270 & 19113 & 0.673 & 28.72\\
        \bottomrule
    \end{tabular}
    \caption{Analysis of variance for world groups and goal groups in the matrix spaces from \figref{fig:teacher student Q} according to \ceqref{eq:within_group_variance} - \ceqref{eq:fvalue}.}

    \label{tab:variances_teacherstudentQ}
\end{table}

\clearpage

\subsection*{Hyperparameters}
In table \ref{tab:hyperparams}, we list the hyperparameters used in our model and their values for different simulations. As we aimed to study the emergent language structure rather than achieve the best performance, we avoided performing a costly hyperparameter search. Instead, values were chosen that lead to reasonable training times and performance. Example simulations have given us good reason to expect stability of the major findings across values of the message length $K$, the learning rate $\alpha_{\textrm{Adam}}$ and also $\gamma$ and $\kappa$, which relate to the student loss function. 

We separate the set of hyperparameters into two groups: task setup and teacher learning (upper half) and language training and student evaluation (lower half). The hyperparameters in the upper half have no significant impact on the communication protocol developed as they are relevant only for teacher learning of the navigation task. 
\begin{table}[h!]
    \centering
    \begin{tabular}{l c c c}
        \toprule
        \textbf{Hyperparameter} & \textbf{usage/meaning} & \textbf{value} & \textbf {figures used} \\
        \midrule
        $n$ & grid-world dimension (including outside walls) & 6 & all \\
        \midrule
        $\tilde{n}$ & grid-world dimension (without outside walls) & 4 & all \\
        \midrule
        $\gamma_{\textrm{Bellman}}$ & temporal discount in teacher Q-learning & 0.99 & all \\
        \midrule
        $R_{\textrm{goal}}$ & goal reward in teacher Q-learning & 2 & all \\
        \midrule
        $R_{\textrm{wall}}$ & wall reward in teacher Q-learning & -0.5 & all \\
        \midrule
        $R_{\textrm{step}}$ & step reward in teacher Q-learning & -0.1 & all \\
        \midrule
        $L$ & short-term memory size in teacher Q-learning & 50 & all \\
        \midrule
        \midrule
        $K$ & message length & 5 & all \\
        \midrule
        $\alpha_{\textrm{Adam}}$ & learning rate in language training & $5\times10^{-4}$ & all \\
        \midrule
        $N_{\textrm{epochs}}$ & epoch number in language training & 1000 & all \\
        \midrule
        $\gamma$ & language training loss weighting & $\frac{1}{20}\sqrt{\frac{4\tilde{n}^{2}}{K}}$ & all \\
        \midrule
        $\zeta$ & language training loss weighting & 5 & all except \figref{fig:autoenc loss plots} \\
        
         &  & 1,2,5,10 & \figref{fig:autoenc loss plots} \\
         \midrule
        $\kappa$ & language training loss weighting & $\frac{1}{500}$ & all \\
        \midrule
        $k$ & allowed steps per task in student evaluation & $2 k_{\textrm{opt}}$ & all \\

        \bottomrule
    \end{tabular}
    \caption{The hyperparameters we used in our model with a brief description as to their usage and their values in the simulations used for creating the different figures. The parameters can roughly be separated into two blocks - the upper block are parameters for grid-world creation and teacher learning of the navigation tasks, whereas the lower block contains parameters for language creation and student evaluation.}
    \label{tab:hyperparams}
\end{table}
\end{document}